\documentclass[10pt,twocolumn,letterpaper]{article}

\usepackage[pagenumbers]{cvpr} 

\definecolor{cvprblue}{rgb}{0.21,0.49,0.74}
\definecolor{mycolor_green}{HTML}{E6F8E0} 
\definecolor{mycolor_gray}{HTML}{ECECEC}
\definecolor{pearDark}{HTML}{2980B9}

\definecolor{T_yellow}{HTML}{F3E060}   
\definecolor{G_teal}{HTML}{4FB6B8}     
\definecolor{R_purple}{HTML}{9C7BD8}   

\definecolor{softOrange}{HTML}{F6B36A}   
\definecolor{paleOrange}{HTML}{FCE8D5}  
\definecolor{brightOrange}{HTML}{FAE3B3}   
\definecolor{lightPeach}{HTML}{F8E0C0}     

\definecolor{brightOrange}{HTML}{F8B878}  
\definecolor{lightPeach}{HTML}{FCE4BD}   

\newcommand{\mixbar}[3]{%
  \begingroup
  \setlength{\fboxsep}{2pt}
  \colorbox{mycolor_gray}{
    \textcolor{G_teal!70!white}{\rule{#2cm}{0.18cm}}%
    \textcolor{T_yellow!75!white}{\rule{#1cm}{0.18cm}}%
    \textcolor{R_purple!70!white}{\rule{#3cm}{0.18cm}}%
  }%
  \endgroup
}

\usepackage{url}            
\usepackage{amsfonts}       
\usepackage{nicefrac}       
\usepackage{microtype}      
\usepackage{xcolor}         
\usepackage{booktabs}       
\usepackage{colortbl}
\usepackage{amsmath}
\usepackage{multirow}
\usepackage{graphicx}   
\usepackage{array} 
\usepackage{amsmath}
\usepackage{adjustbox}
\usepackage{makecell} 
\definecolor{citecolor}{HTML}{2980b9}
\definecolor{linkcolor}{HTML}{c0392b}
\usepackage[pagebackref,breaklinks,colorlinks,citecolor=citecolor,linkcolor=linkcolor]{hyperref}

\def\paperID{***} 
\def\confName{CVPR}
\def\confYear{2026}

\title{\vspace{-0.6cm}\hspace{-1.5cm}\includegraphics[height=2.5em]{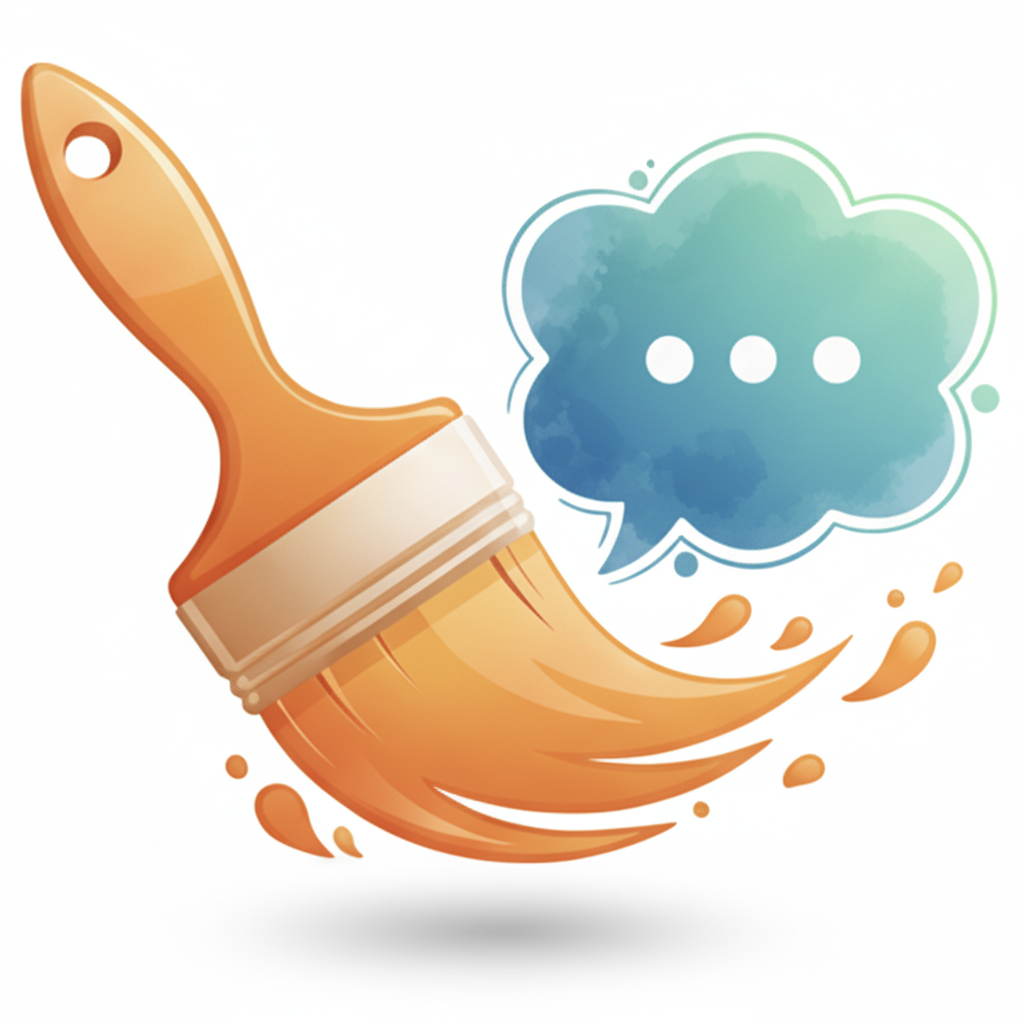} \textit{Thinking-while-Generating:}\\Interleaving Textual Reasoning throughout Visual Generation}

\author{Ziyu Guo$^{*1}$, Renrui Zhang$^{\dagger*2}$, Hongyu Li$^{*3}$, Manyuan Zhang$^{\dagger3}$, Xinyan Chen$^{2}$\vspace{0.07cm}\\Sifan Wang, 
Yan Feng$^{3}$, Peng Pei$^{3}$, Pheng-Ann Heng$^{1}$\vspace{0.35cm}\\
CUHK $^1$IMIXR\hspace{0.1cm} \&\hspace{0.1cm} $^2$MMLab \quad
$^3$Meituan\vspace{0.3cm}\\
\centerline{Project Page: \url{https://think-while-gen.github.io}}
}

\begin{document}

\maketitle

\begin{abstract}
Recent advances in visual generation have increasingly explored the integration of reasoning capabilities. They incorporate textual reasoning, i.e., \textit{think}, either before (as pre-planning) or after (as post-refinement) the generation process, yet they lack on-the-fly multimodal interaction during the generation itself.
In this preliminary study, we introduce \textbf{Thinking-while-Generating} (\textsc{TwiG}), \textbf{the first} interleaved framework that enables co-evolving textual reasoning throughout the visual generation process. As visual content is progressively generating, textual reasoning is interleaved to both guide upcoming local regions and reflect on previously synthesized ones. This dynamic interplay produces more context-aware and semantically rich visual outputs.
To unveil the potential of this framework, we investigate three candidate strategies, zero-shot prompting, supervised fine-tuning (SFT) on our curated \textsc{TwiG}-50K dataset, and reinforcement learning (RL) via a customized \textsc{TwiG}-GRPO strategy, each offering unique insights into the dynamics of interleaved reasoning.
We hope this work inspires further research into interleaving textual reasoning for enhanced visual generation. Code will be released at: \url{https://github.com/ZiyuGuo99/Thinking-while-Generating}.
\end{abstract}

\renewcommand{\thefootnote}{\fnsymbol{footnote}}

\footnotetext{$^*$Equal Contribution\hspace{0.2cm} $^\dagger$Project Lead}

\section{Introduction}
\begin{figure}[t]
\centering
    \includegraphics[width=0.95\linewidth]{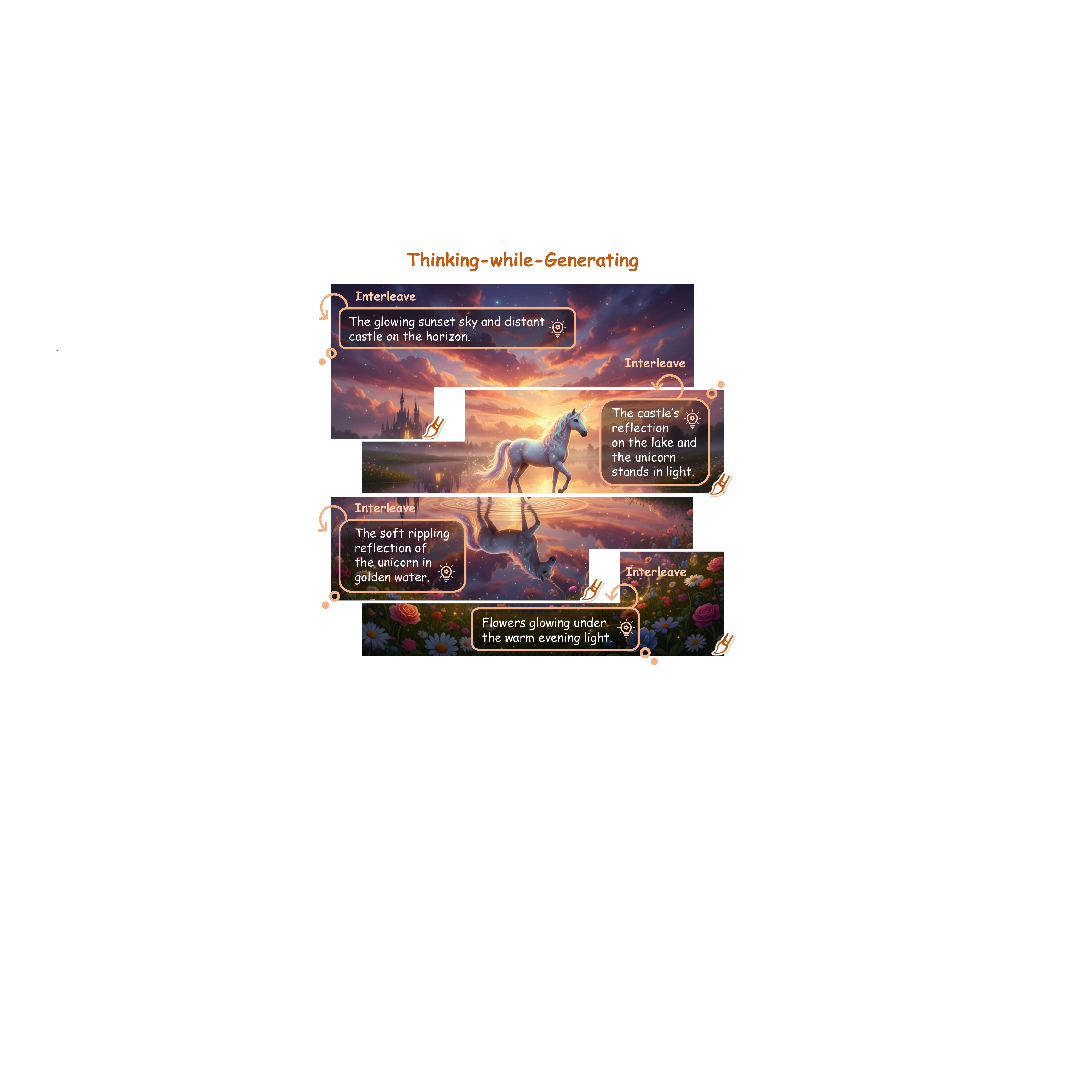}
    \caption{\textbf{Interleaving Textual Reasoning throughout Visual Generation.} Inspired by the image-interleaved reasoning in textual responses~\cite{su2025thinking,zheng2025deepeyes,openai_o3,chen2025mint}, we reverse the modality flow and weave textual thoughts into the unfolding canvas, delivering on-the-fly guidance and reflection throughout synthesis.}
    \label{fig:teaser}
\end{figure}

\begin{figure*}[t]
    \includegraphics[width=\linewidth]{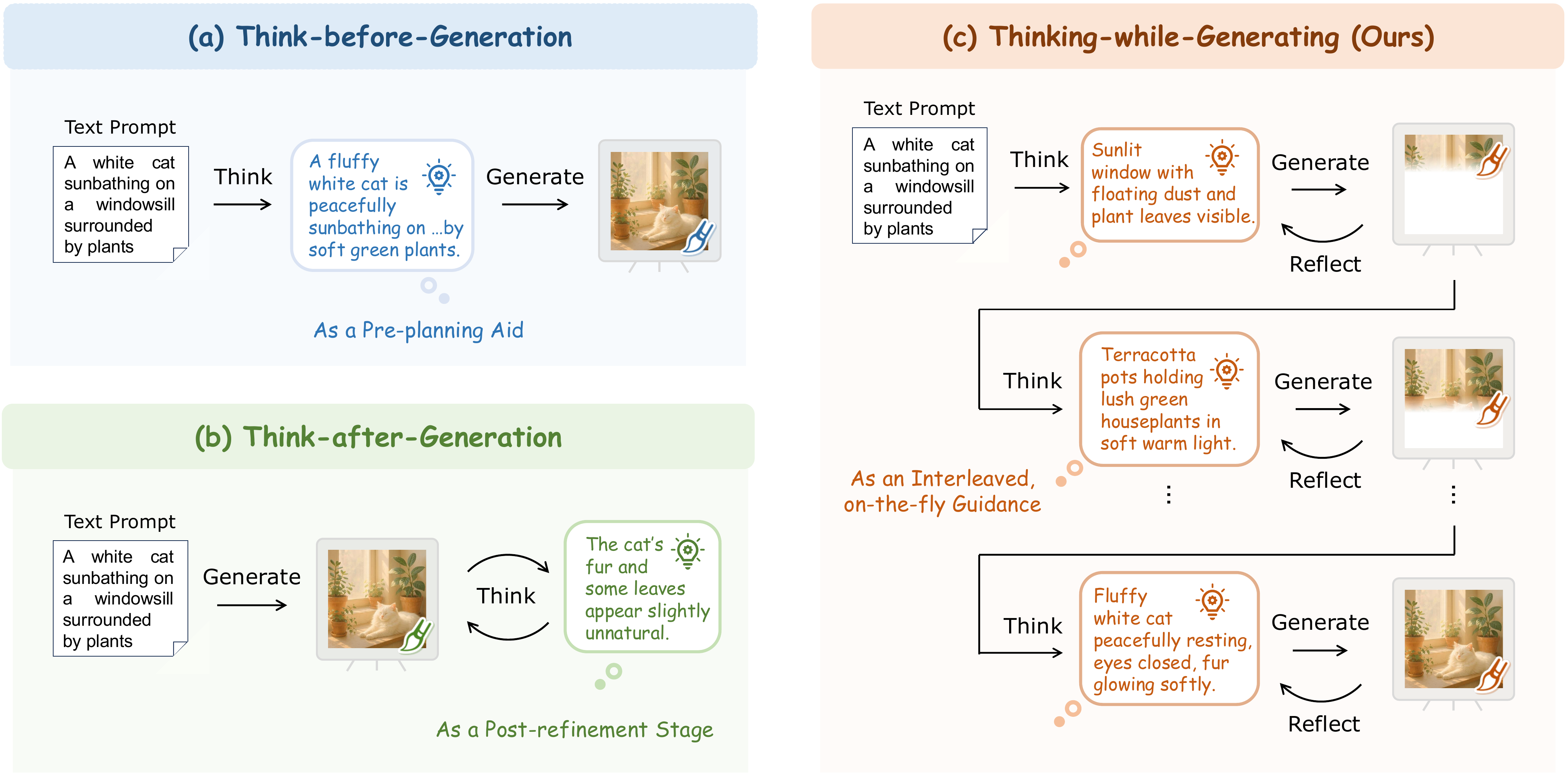}
    \caption{\textbf{Comparison of \textit{Where} the Textual Reasoning is Applied in Visual Generation:} (a) \textit{Think-before-Generation}~\cite{jiang2025t2i,fang2025got,liao2025imagegen} injects a pre-planning thought prior to the synthesis, limiting fine-grained control and later correction; (b) \textit{Think-after-Generation}~\cite{guo2025can,qin2025uni,li2025reflect} verifies and revise the image once it is complete, lacking nuanced, timely adjustment with extra inference cost; (c) Our \textit{Thinking-while-Generating} interleaves thoughts and reflections throughout the synthesis, providing on-the-fly. co-evolving guidance.}
    \label{fig:intro}
\vspace{0.3cm}
\end{figure*}

Visual generation have developed rapidly with diffusion~\cite{ramesh2022hierarchical, saharia2022imagen, rombach2022ldm} and autoregressive~\cite{xie2024show,team2024chameleon,chen2025janus} models, enabling high-fidelity synthesis across diverse domains~\cite{poole2022dreamfusion,zhang2023adding,openai2025sora2}. Despite impressive visual quality, today’s generators often struggle with long-horizon composition, multi-entity relations, and adherence to nuanced textual instructions. Starting from \textit{`Generation with CoT'}~\cite{guo2025can,tong2025delving}, a growing line of work explores \emph{reasoning} as a remedy, typically injecting chain-of-thoughts in the language modality to assist visual synthesis.\vspace{0.1cm}

Existing CoT-based approaches can be grouped by \emph{where} the textual reasoning is applied, as compared in Figure~\ref{fig:intro}:\vspace{0.1cm}
\begin{itemize}
    \item \textit{Think before generation as a pre-planning aid.} Methods~\cite{jiang2025t2i,fang2025got,liao2025imagegen} first produce a structured or free-form plan, e.g., detailed captions, scene layouts, or object attributes and relations, and then condition the image generator on this plan. This improves global coherence and entity placement, but the plan is fixed once generation begins, limiting nuanced guidance and mid-course correction.\vspace{0.1cm}

    \item \textit{Think after generation as a post-refinement stage.} Methods~\cite{guo2025can,zhuo2025reflection,li2025reflect} synthesize the entire image first, and then elicit textual feedback via self-critique or external verifiers, iteratively revising the visual errors. These approaches help with local fixes and attribute binding, but reasoning is only loosely coupled to the synthesis trajectory without fine-grained, timely revision and, importantly, incur additional, costly extra inference rounds.
\end{itemize}

Given these limitations in visual generation, we note a complementary trend in visual understanding: recent large multimodal models (LMMs)~\cite{li2025imagine,duan2025codeplot,zheng2025deepeyes,gao2025interleaved,zhang2024mavis} perform image-text interleaved reasoning, adaptively weaving intermediate visual evidence (e.g., detected objects, zoomed-in regions, or tagged images) into textual CoTs to improve interpretation and analysis. Inspired by this paradigm, we pose a natural question: as illustrated in Figure~\ref{fig:teaser}, \emph{Can we invert the flow and interleave text into the intermediate visual generation process, providing on-the-fly, co-evolving reasoning that guides synthesis as it unfolds?}

In this preliminary study, we present \textit{the first} interleaved framework for visual generation that keeps textual reasoning in the loop, termed as \textbf{\textit{Thinking-while-Generating}} (\textsc{TwiG}), as compared in Figure~\ref{fig:intro} (c). As our approach is compatible with multiple models and task settings, for clarity and future extensibility, we adopt the unified understanding-generation LMM (ULM)~\cite{wu2024janus,xie2024show,zhou2024transfusion,chen2025janus} with autoregressive generation paradigms, e.g., Janus-Pro~\cite{chen2025janus}, and experiment on text-to-image scenarios in our study.

Given a text prompt, the model first interprets the instruction and plans an optimal interleave schedule, i.e., how many steps to use and how to partition the canvas into local regions for progressive synthesis. While generating each region, the model conducts on-the-fly textual reasoning and grounds its thoughts in the current partial visual state. This interleaved \emph{think} step serves two roles: \textit{(i)} it produces nuanced guidance for the upcoming synthesis, and \textit{(ii)} it critiques and reflects on the previously generated content. In this way, textual reasoning co-evolves with the visual modality, providing detailed, step-by-step directives. The image can be dynamically revised and precisely steered as it unfolds within a single generative trajectory.

We consider three candidate routes for \emph{Thinking-while-Generating}, and investigate which, if any, proves effective:

\begin{itemize}
    \item \textit{Can zero-shot prompting alone achieve the goal?}
    We craft interleave-aware prompts to directly elicit global plans and reasoning thoughts. This route reveals the latent capacity of ULMs to self-organize interleaved reasoning without parameter updates, but can suffer instability.
    \vspace{0.1cm}
    
    \item \textit{Does supervised fine-tuning (SFT) benefit the performance?}
    We categorize the understanding and generation process into nine subtasks, and curate a dataset, \textsc{TwiG}-50K, for fine-tuning ULM, aiming to improve instruction adherence and reduce visual hallucination.
    \vspace{0.1cm}
    
    \item \textit{Will reinforcement learning (RL) further unlock its potential?}
    We optimize the interleaved reasoning policy of ULM via a customized GRPO~\cite{shao2024deepseekmath} algorithm, \textsc{TwiG}-GRPO, to push the performance boundary, investigating different RL approaches and reward designs.
    \vspace{0.1cm}
\end{itemize}

Our experiments indicate that the ULM itself exhibits strong zero-shot capability for \emph{Thinking-while-Generating}. With carefully designed prompts, it substantially improves Janus-Pro on T2I-CompBench(++)~\cite{huang2023t2i,huang2025t2i} without additional training. 
Building on this, SFT with \textsc{TwiG}-50K provides further modest yet consistent gains, leading to more stable behavior compared with the zero-shot baseline. 
Finally, optimization with our \textsc{TwiG}-GRPO algorithm yields considerable improvements, underscoring the value of RL for deciding \emph{when} to think, \emph{what} to say, and \emph{how} to refine. Taken together, these findings, though preliminary, are informative: they demonstrate the feasibility of interleaving textual reasoning during generation, and highlight this direction as a promising avenue for advancing visual synthesis.\vspace{0.1cm}

It is worth noting that two relevant concurrent works, IRG~\cite{huang2025interleaving} and Uni-CoT~\cite{qin2025uni}, attempt to `interleave' reasoning with generation, but still treat the visual systhesis process as a monolithic block, like a combination of \textit{think-before-generation} and \textit{think-after-generation}. 
They are well-performed with unique insights, but not truly interleaving reasoning within the generative process itself, limiting the granularity and controllability.

\section{\textit{Thinking-while-Generating}}
In Section~\ref{s3.1}, we first introduce the design scope and applicability of \emph{Thinking-while-Generating}. Then, in Section~\ref{s3.2}, we present its overall pipeline and core components of the framework in detail.

\begin{figure*}[t]
    \hspace{0.3cm}\includegraphics[width=0.95\linewidth]{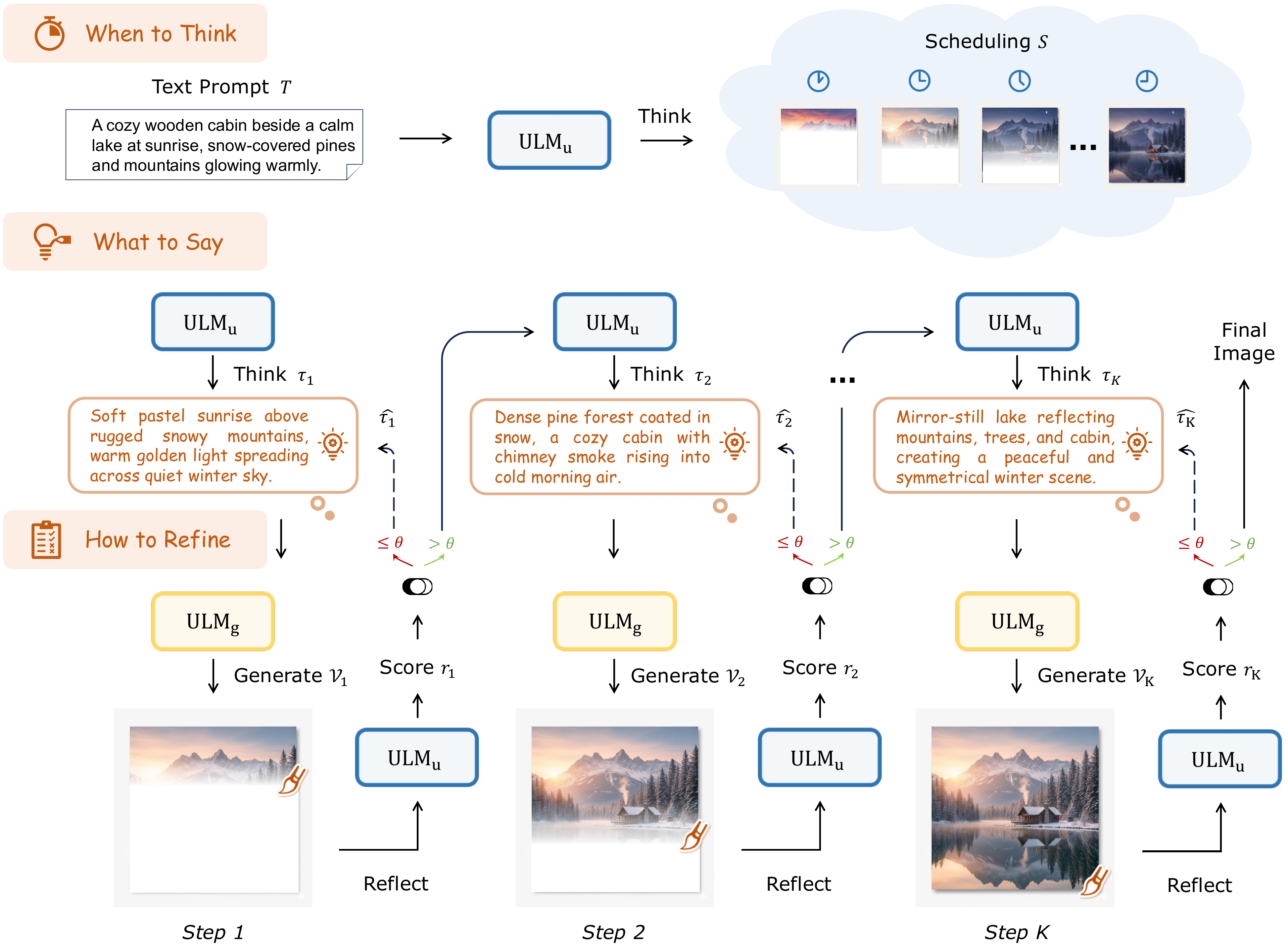}
    \caption{\textbf{Overall Pipeline of \textit{Thinking-while-Generating}.} The framework comprises three components: \textit{When to Think} for globally determining the interleaved generation schedule; \textit{What to Say} for producing the step-by-step textual thought as fine-grained guidance; and \textit{How to Refine} for a region-level reflection on the current canvas with optional corrective updates. $\mathrm{ULM}_{u}$ and $\mathrm{ULM}_{g}$ denote to apply a single ULM for understanding and generation, respectively.}
    \label{fig:pipeline}
\end{figure*}

\subsection{Scope and Applicability}
\label{s3.1}

Aiming for generalization and extensibility, \emph{Thinking-while-Generating} is conceptually compatible with diverse settings along the following three axes:\vspace{0.1cm}

\begin{itemize}
    \item \textbf{System Architecture.} The framework can be instantiated either as (i) a pipeline that couples a text-to-image model~\cite{betker2023improving,esser2024scaling,ramesh2022hierarchical} with an LMM~\cite{yang2025qwen3,achiam2023gpt,liu2023llava}, where the LMM specializes in producing interleaved reasoning for the text-to-image outputs; or (ii) a ULM~\cite{chen2025janus,xie2024show,zhou2024transfusion} that performs textual reasoning and visual generation within a single backbone.\vspace{0.1cm}

    \item \textbf{Generation Paradigm.} The framework is applicable for visual generation with diffusion~\cite{feng2022training, liu2022compositional,rombach2022high}, discrete diffusion~\cite{xie2024show,chang2022maskgit,xie2025show}, and autoregressive models~\cite{chen2025janus,team2024chameleon,sun2024autoregressive}. For continuous diffusion models, textual thoughts are interleaved at selected denoising steps; for discrete diffusion and autoregressive models, thoughts are inserted between segments of visual tokens to guide upcoming spans.\vspace{0.1cm}

    \item \textbf{Task Scenarios.} The framework applies beyond T2I, e.g., image-to-image~\cite{brooks2023instructpix2pix,hertz2022prompt,zhang2023adding}, text-to-video~\cite{wan2025wan,GoogleDeepMind2025Veo3,guo2025video}, text-to-3D~\cite{poole2022dreamfusion,lin2023magic3d,guo2023point}, and related generative tasks: as long as an LMM (or ULM) can provide reasoning thoughts for the target modality, they can be interleaved to steer generation.\vspace{0.1cm}
\end{itemize}

As a preliminary study, we adopt a \emph{single} ULM with an autoregressive generation paradigm (e.g., Janus-Pro~\cite{chen2025janus}) for clarity of exposition, promising headroom, and end-to-end training efficiency. We denote its understanding forward pass by $\mathrm{ULM}_{u}$ and the generation forward pass by $\mathrm{ULM}_{g}$.

\vspace{0.2cm}
\subsection{Framework Overview}
\label{s3.2}

Figure~\ref{fig:pipeline} presents the overall \textit{Thinking-while-Generating} (\textsc{TwiG}) framework, which interleaves textual reasoning with visual generation through three schemes: \emph{when} to think, \emph{what} to say, and \emph{how} to refine.

\paragraph{\emph{When} to Think (Scheduling).} Given an input prompt $T$, $\mathrm{ULM}_u$ first determines an interleaved reasoning schedule, denoted as $\mathcal{S} = \{\mathcal{V}_k\}_{k=1}^{K}$, according to:
\[
\mathcal{S} = \mathrm{ULM}_{u}(T),
\]
where each $\mathcal{V}_k$ denotes a target visual region at which reasoning is applied (e.g., token spans in autoregressive and discrete diffusion models, or timestep windows in continuous diffusion models). This decouples the generation process into smaller, more controllable sub-tasks guided by the interleaved textual reasoning.
Scheduling can be static (fixed $K$, uniform spacing) or adaptive (variable $K$, content-dependent $\mathcal{V}_k$). 
In Section~\ref{s4.1}, we investigate different schedules and find that a static schedule with $K=3$ performs the best, based on the heuristic that most images consist of three semantic components: upper background, central content, and lower background. Additionally, current capabilities of $\mathrm{ULM}_u$ are limited in reliably generating well-structured adaptive schedules, which remains a future work.

\begin{figure}[t]
\centering
    \includegraphics[width=0.95\linewidth]{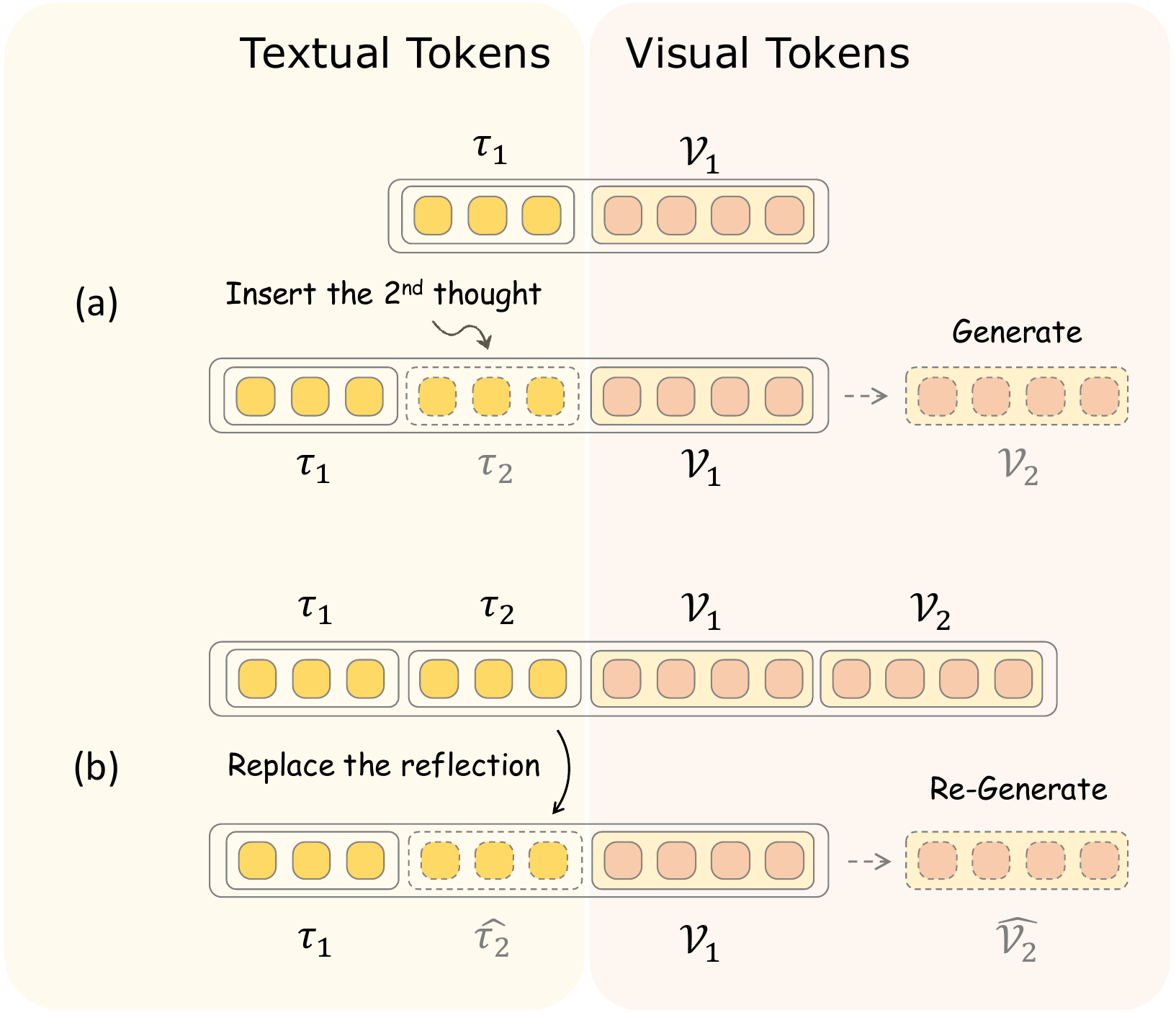}
    \caption{\textbf{Illustration of Interleaved Token Sequence:} (a) In \textit{What to Say}, the textual pre-context extends from $\{\tau_j\}_{j<k}$ to $\{\tau_j\}_{j\le k}$ ($K = 2$), guiding the generation of the next $\mathcal{V}_k$ while leaving the earlier $\{\mathcal{V}_j\}_{j<k}$ untouched; (b) In \textit{How to Refine}, the thought ${\tau}_{k}$ is revised to $\hat{\tau}_{k}$, and only the local region $\hat{\mathcal{V}}_k$ is re-generated to replace $\mathcal{V}_k$. Neither operation requires the ULM to possess image-to-image capabilities, and both preserve a single text-to-image generation trajectory without launching a fresh pass or full re-generation.}
    \label{fig:motivation}
\end{figure}

\paragraph{\emph{What} to Say (Reasoning Content).}
At each scheduled reasoning point, $\mathrm{ULM}_{u}$ provides a textual thought $\tau_k$ intended to guide the generation of the visual region $\mathcal{V}_k$. 
This thought serves as a localized sub-prompt exclusively targeted at $\mathcal{V}_k$, offering finer-grained guidance and alignment than prior \emph{think-before-generation} approaches.
The generation of $\tau_k$ is conditioned on three elements, i.e., the input prompt $T$, the previous thoughts $\{\tau_j\}_{j<k}$, and the visual content generated for prior regions $\{\mathcal{V}_j\}_{j<k}$, formulated as:
\[
\tau_k = \mathrm{ULM}_{u}(T,\ \{\tau_j\}_{j<k},\ \{\mathcal{V}_j\}_{j<k}).
\]
This allows $\tau_k$ to incorporate accumulated contextual information and to plan appropriately for the next visual segment.
Subsequently, $\mathrm{ULM}_{g}$ synthesizes the target region $\mathcal{V}_k$, conditioned on all reasoning thoughts and the visual content produced up to by:
\[
\mathcal{V}_k = \mathrm{ULM}_{g}(\{\tau_j\}_{j \leq k},\ \{\mathcal{V}_j\}_{j<k}).
\]
It is important to note that \textbf{\textit{$\mathrm{ULM}_{g}$ is only required to possess text-to-image capabilities, no need for image-to-image functionality}}. This is because the visual context $\{\mathcal{V}_j\}_{j<k}$ is not provided as image input to the model. Instead, we directly extend the textual pre-context from $\{\tau_j\}_{j<k}$ to $\{\tau_j\}_{j \leq k}$ at the beginning of the token sequence, while preserving the generated visual content $\{\mathcal{V}_j\}_{j<k}$ unchanged at the end of the sequence.
This modification preserves the autoregressive generation process within a single trajectory, without introducing discontinuities or new generation rounds, as illustrated in Figure~\ref{fig:motivation} (a).

\paragraph{\emph{How} to Refine (Reflection).}
After generating each visual region $\mathcal{V}_k$, we allow $\mathrm{ULM}_{u}$ to perform an immediate, region-level revision step that couples visual critique and an optional correction process.
This enables finer-grained corrections while significantly reducing computational cost compared to prior \emph{think-after-generation} approaches that conduct global post-revision.
Before producing the next reasoning thought $\tau_{k+1}$, $\mathrm{ULM}_{u}$ first generates a reflection tuple $c_k = (r_k, \hat{\tau}_{k})$, given all the generated textual and visual contents as:
\[
c_k = \mathrm{ULM}_{u}(T,\ \{\tau_j\}_{j \leq k},\ \{\mathcal{V}_j\}_{j \leq k}),
\]
where $r_k \in [0,100]$ is an integer representing the critic score assigned to the current region $\mathcal{V}_k$, and $\hat{\tau}_{k}$ is a revised sub-caption intended for potential correction.
The score $r_k$ evaluates the semantic alignment and visual coherence of $\mathcal{V}_k$ with respect to its guiding prompt $\tau_k$. If $r_k$ exceeds a predefined threshold $\theta$, the model proceeds directly to generate the next reasoning thought without revision. Otherwise, a local reflection is triggered to refine only the current sub-region, guided by $\hat{\tau}_{k}$, as defined by:
\[
\hat{\mathcal{V}}_k = \mathrm{ULM}_{g}(\{\tau_j\}_{j < k},\ \hat{\tau}_{k},\ \{\mathcal{V}_j\}_{j<k}).
\]
This localized corrective mechanism mitigates the accumulation of visual misalignments with timely revision. Likewise, as presented in Figure~\ref{fig:motivation} (b), we directly update the textual pre-context from ${\tau}_{k}$ to the revised $\hat{\tau}_{k}$, and re-generate only the local part $\hat{\mathcal{V}}_k$ to replace $\mathcal{V}_k$ at the end of the token sequence, which also preserves a single trajectory without requiring the costly full re-generation.
\vspace{0.1cm}

In sum, \emph{Thinking-while-Generating} (i) first schedules a number $K$ of interleaved reasoning points (\emph{when}); then for each $k=1,\dots,K$, (ii) produces a textual thought that locally steers the next visual update (\emph{what}); and (iii) performs a region-level reflection with optional correction (\emph{how}). The loop of (ii) and (iii) preserves a single generative trajectory, enabling on-the-fly guidance and precise local revision.

\section{Implementation Exploration}

In this section, we implement three candidate approaches for \emph{Thinking-while-Generating}: zero-shot prompting (\ref{s4.1}), supervised fine-tuning (\ref{s4.2}), and reinforcement learning (\ref{s4.3}). We present experimental results that highlight their respective strengths. Please refer to detailed experimental settings and visualizations in the Supplementary Material.

\begin{table}[t]
\centering
\caption{{\bf Zero-shot Experiments} of \textit{Thinking-while-Generating} on T2I-CompBench~\cite{huang2023t2i}. We denote our zero-shot model as \textsc{TwiG-ZS}, and mark the improvement over the baseline, Janus-Pro-7B~\cite{chen2025janus}. Panels \textit{(a)}, \textit{(b)}, \textit{(c)}, and \textit{(d)} present four ablation studies.}
\label{tab:zs_main_ablation}
\begin{adjustbox}{width=\linewidth}
\begin{tabular}{l@{\hspace{0.45cm}}cccccc}
\toprule
\multirow{2}{*}{\bf Setting} & \multicolumn{3}{c}{\bf Attribute Binding} & \multicolumn{2}{c}{\bf Object Relationship} & \multirow{2}{*}{\bf Complex$\uparrow$}\\
\cmidrule(lr){2-4}\cmidrule(lr){5-6}
& {\bf Color$\uparrow$} & {\bf Shape$\uparrow$} & {\bf Texture$\uparrow$} & {\bf Spatial$\uparrow$} & {\bf Non-Spatial$\uparrow$} & \\
\midrule
\multicolumn{7}{c}{\textit{v.s. Baseline}}\\
\cmidrule{1-7}
Janus-Pro-7B~\cite{chen2025janus} & 63.59 & 35.28 & 49.36 & 20.61 & 30.85 & 35.59  \\
\textsc{TwiG-ZS} & 73.11 & 41.55 & 64.77 & 21.98 & 30.90 & 48.16 \\
\textcolor{blue}{\textit{Improve}} & \textcolor{blue}{\textit{+9.52 }} & \textcolor{blue}{\textit{+6.27 }} & \textcolor{blue}{\textit{+15.41 }} & \textcolor{blue}{\textit{+1.37 }} & \textcolor{blue}{\textit{+0.05 }} & \textcolor{blue}{\textit{+12.57}} \\
\midrule
\multicolumn{7}{c}{\textit{(a) Where the Textual Reasoning is Applied}}\\
\cmidrule{1-7}
\textit{Think-before-Gen.}  & 65.12 & 36.20 & 51.05 & 20.88 & 30.82 & 41.75 \\
\textit{Think-after-Gen.} & 64.72 & 37.95 & 50.62 & 21.05 & 30.87 & 42.28 \\
\textit{Thinking-while-Gen.} & 73.11 & 41.55 & 64.77 & 21.98 & 30.90 & 48.16 \\
\midrule
\multicolumn{7}{c}{\textit{(b) Interleaved Reasoning Step}}\\
\cmidrule{1-7}
$K=2$ & 72.79 & 42.26 & 64.64 & 21.97 & 30.89 & 49.71 \\
$K=3$ & 73.11 & 41.55 & 64.77 & 21.98 & 30.90 & 48.16 \\
$K=4$ & 72.95 & 41.90 & 64.70 & 22.03 & 31.10 & 48.90 \\
\midrule
\multicolumn{7}{c}{\textit{(c) How to Partition $\mathcal{V}_k$} in Space}\\
\cmidrule{1-7}
Uniform Spacing & 73.11 & 41.55 & 64.77 & 21.98 & 30.90 & 48.16 \\
Adaptive Spacing & 72.43 & 40.88 & 63.92 & 21.67 & 30.88 & 47.39 \\
\midrule
\multicolumn{7}{c}{\textit{(d) Whether to Perform Reflection}}\\
\cmidrule{1-7}
w/o Reflection & 73.11 & 41.55 & 64.77 & 21.98 & 30.90 & 48.16 \\
1-round Reflection & 73.90 & 46.02 & 66.10 & 24.50 & 30.81 & 51.97 \\
2-round Reflection & 73.68 & 45.72 & 66.02 & 24.42 & 30.88 & 51.65 \\
\bottomrule
\end{tabular}
\end{adjustbox}
\end{table}

\subsection{Zero-shot Prompting}
\label{s4.1}

\paragraph{Prompt Customization.}
To elicit satisfactory zero-shot \emph{Thinking-while-Generating}, we meticulously design a series of interleave-aware prompts for ULM, corresponding to the three components described in Section~\ref{s3.2}. Please refer to the final prompt templates in the Supplementary Material.

\begin{itemize}
    \item For \textit{when} to think, we prompt the model to adopt a global view, sketching the image’s high-level semantics and structure step by step from the input prompt. For an adaptive schedule, we additionally prompt the model to output the relative ratios of visual parts across the canvas.\vspace{0.1cm}
    
    \item For \textit{what} to say, we guide the model to focus strictly on the local region currently being generated while maintaining coherence with previously generated visual and textual context. We discourage any spatial-anchor tokens; the model should produce only the descriptive content.\vspace{0.1cm}
    
    \item For \textit{how} to refine, we prompt the model to provide a critic score evaluating along five criteria (color accuracy, object completeness, detail richness, spatial relationships, and visual coherence), ensuring a consistent standard across cases. The template enforces that any revision is local and does not contradict validated prior regions.\vspace{0.1cm}
\end{itemize}

\vspace{-0.2cm}
\paragraph{Experiments and Analysis.}
In Table~\ref{tab:zs_main_ablation} (top), we present the performance of our zero-shot model, \textsc{TwiG-ZS}.
We observe that our carefully designed prompts yield surprisingly strong improvements over the baseline, significantly surpassing Janus-Pro-7B~\cite{chen2025janus} across multiple dimensions. This highlights the potential of our framework and its natural applicability within current ULMs, making the zero-shot variant a strong foundation for subsequent SFT and RL.
By default, we adopt an interleaved schedule with $K=3$ and uniform spacing, and permit at most one round of reflection. We conduct four ablations:\vspace{0.1cm}
\begin{itemize}
    \item \textbf{Ablation \textit{(a):}} \textit{Thinking-while-Generating} versus \textit{Think-before/after-Generation} under identical zero-shot settings. Interleaving provides nuanced, on-the-fly guidance rather than only pre-planning or post-refinement, and consistently outperforms the alternatives.\vspace{0.1cm}

    \item \textbf{Ablation \textit{(b):}} Number of interleaved reasoning steps under a uniform schedule. We find $K=3$ is optimal, aligning with the heuristic that many images decompose into three semantic components: upper background, central content, and lower background.\vspace{0.1cm}

    \item \textbf{Ablation \textit{(c):}} Adaptive scheduling of interleaved spacing. Despite exploring multiple prompting strategies, current ULMs struggle to reliably follow such instructions, leading to unstable or poorly structured adaptive schedules.\vspace{0.1cm}

    \item \textbf{Ablation \textit{(d):}} Effectiveness of reflection during reasoning. A single reflection round corrects misalignments and improves performance across aspects; however, conducting two rounds brings no further gains, likely limited by the critique–and–revision capacity of zero-shot ULMs.
\end{itemize}

\begin{table}[t]
\centering
\caption{{\bf SFT Experiments} of \textit{Thinking-while-Generating} on T2I-CompBench~\cite{huang2023t2i}. We denote our fine-tuned model as \textsc{TwiG-SFT}, and mark the improvement over \textsc{TwiG-ZS}. Panel \textit{(a)} ablates the varying proportions of thinking (\textcolor{G_teal!70!white}{\textbf{\textit{T}}}), generation (\textcolor{T_yellow!}{\textbf{\textit{G}}}), and reflection (\textcolor{R_purple!70!white}{\textbf{\textit{R}}}) data in \textsc{TwiG}-50K. Panel \textit{(b)} reports the standard deviation (\textit{Std}) across random seeds to assess stability.}
\label{tab:sft_mix_nexttosetting}
\begin{adjustbox}{width=\linewidth}
\begin{tabular}{l@{\hspace{0.3cm}}lcccccc}
\toprule
\multirow{2}{*}{\bf Model / Setting} & \makecell[c]{\bf Data} &
\multicolumn{3}{c}{\bf Attribute Binding} &
\multicolumn{2}{c}{\bf Object Relationship} &
\multirow{2}{*}{\bf Complex$\uparrow$} \\
\cmidrule(lr){3-5}\cmidrule(lr){6-7}
&\makecell[c]{\textcolor{G_teal!70!white}{\textbf{\textit{T}}} / 
\textcolor{T_yellow!}{\textbf{\textit{G}}} / 
\textcolor{R_purple!70!white}{\textbf{\textit{R}}}} 
& {\bf Color$\uparrow$} & {\bf Shape$\uparrow$} & {\bf Texture$\uparrow$} 
& {\bf Spatial$\uparrow$} & {\bf Non-Spatial$\uparrow$} & \\

\midrule
\multicolumn{8}{c}{\textit{v.s. Baseline}}\\
\cmidrule{1-8}
Janus-Pro-7B~\cite{chen2025janus} & \makecell[c]{--} 
& 63.59 & 35.28 & 49.36 & 20.61 & 30.85 & 35.59  \\
\textsc{TwiG-ZS}  & \makecell[c]{--} 
& 73.11 & 41.55 & 64.77 & 21.98 & 30.90 & 48.16 \\
\textsc{TwiG-SFT} & \mixbar{0.75}{0.75}{0.0} 
& 74.58 & 52.42 & 67.95 & 27.02 & 31.24 & 53.41 \\
\textcolor{blue}{\textit{Improve}} 
& \makecell[c]{--} 
& \textcolor{blue}{\textit{+1.47}}
& \textcolor{blue}{\textit{+10.87}}
& \textcolor{blue}{\textit{+3.18}}
& \textcolor{blue}{\textit{+5.04}}
& \textcolor{blue}{\textit{+0.34}}
& \textcolor{blue}{\textit{+5.25}} \\
\midrule
\multicolumn{8}{c}{\textit{(a) Effect of Training Data Composition}}\\
\midrule
Think-heavy        & \mixbar{0.5}{1.0}{0.0}   
& 73.38 & 50.92 & 66.47 & 26.08 & 30.97 & 51.86 \\
Gen-heavy          & \mixbar{1.0}{0.5}{0.0}   
& 74.12 & 51.77 & 67.28 & 26.58 & 31.09 & 52.83 \\
Think-Gen-equal    & \mixbar{0.75}{0.75}{0.0} 
& 74.58 & 52.42 & 67.95 & 27.02 & 31.24 & 53.41 \\
Reflect-lite       & \mixbar{0.64}{0.64}{0.21} 
& 72.76 & 49.75 & 65.93 & 26.36 & 30.92 & 51.17 \\
Reflect-heavy      & \mixbar{0.5}{0.5}{0.5}   
& 71.88 & 48.98 & 65.05 & 25.62 & 30.84 & 50.27 \\
\midrule
\multicolumn{8}{c}{\textit{(b) Stability across 5 Random Seeds}}\\
\midrule
\textsc{TwiG-ZS} \textit{Std}$\downarrow$ & \makecell[c]{--}   
& 0.82 & 0.70 & 0.76 & 0.45 & 0.38 & 0.91 \\
\textsc{TwiG-SFT} \textit{Std}$\downarrow$ & \mixbar{0.75}{0.75}{0.0} 
& 0.65 & 0.59 & 0.61 & 0.40 & 0.36 & 0.80 \\ 
\bottomrule
\end{tabular}
\end{adjustbox}
\end{table}

\vspace{0.1cm}
\subsection{Supervised Fine-tuning}
\label{s4.2}
\vspace{0.1cm}

\paragraph{SFT Task Formulation.}
Building on the zero-shot baseline, we investigate whether SFT can enhance the capabilities. We decompose the \emph{Thinking-while-Generating} process into nine supervised tasks that mirror the inference loop, using a fixed number of three reasoning steps. These comprise three thinking targets for $\mathrm{ULM}_{u}$ (upper/central/lower thoughts), three reflection targets for $\mathrm{ULM}_{u}$ (three scores with revised thoughts), and three generation targets for $\mathrm{ULM}_{g}$ (three visual regions). This enables the model to learn structured reasoning, localized reflection, and region-wise generation in an interleaved, context-aware manner.\vspace{-0.2cm}

\begin{table}[t]
\centering
\caption{{\bf RL Experiments} of \textit{Thinking-while-Generating} on T2I-CompBench~\cite{huang2023t2i}. We denote our reinforced model with GRPO~\cite{shao2024deepseekmath_grpo} as \textsc{TwiG-RL}, and mark the improvement over the \textsc{TwiG-SFT}. Panels \textit{(a)} and \textit{(b)} present the results of two ablation studies.}
\label{tab:rl_main_ablation}
\begin{adjustbox}{width=\linewidth}
\begin{tabular}{l@{\hspace{0.45cm}}cccccc}
\toprule
\multirow{2}{*}{\bf Setting} & 
\multicolumn{3}{c}{\bf Attribute Binding} & 
\multicolumn{2}{c}{\bf Object Relationship} & 
\multirow{2}{*}{\bf Complex$\uparrow$}
\\
\cmidrule(lr){2-4}\cmidrule(lr){5-6}
& {\bf Color$\uparrow$} & {\bf Shape$\uparrow$} & {\bf Texture$\uparrow$} 
& {\bf Spatial$\uparrow$} & {\bf Non-Spatial$\uparrow$} & 
\\
\midrule
\multicolumn{7}{c}{\textit{v.s. Baseline}}\\
\cmidrule{1-7}

Janus-Pro-7B~\cite{chen2025janus}
& 63.59 & 35.28 & 49.36 & 20.61 & 30.85 & 35.59 \\

\textsc{TwiG-ZS}
& 73.11 & 41.55 & 64.77 & 21.98 & 30.90 & 48.16 \\

\textsc{TwiG-SFT}
& 74.58 & 52.42 & 67.95 & 27.02 & 31.24 & 53.41 \\

\textsc{TwiG-RL}
& 82.49 & 61.28 & 73.19 & 34.06 & 31.99 & 54.45 \\

\textcolor{blue}{\textit{Improve}}
& \textcolor{blue}{\textit{+7.91}}
& \textcolor{blue}{\textit{+8.86}}
& \textcolor{blue}{\textit{+5.24}}
& \textcolor{blue}{\textit{+7.04}}
& \textcolor{blue}{\textit{+0.75}}
& \textcolor{blue}{\textit{+1.04}}
\\

\midrule

\multicolumn{7}{c}{\textit{(a) \textsc{TwiG}-GRPO Strategy}}\\
\cmidrule{1-7}

$\mathrm{ULM}_g$-GRPO 
& 80.12 & 59.87 & 72.01 & 32.47 & 31.30 & 54.02 \\

$\mathrm{ULM}_u$-GRPO 
& 78.36 & 57.94 & 70.68 & 30.93 & 31.27 & 53.76 \\

\textsc{TwiG}-GRPO & 82.49 & 61.28 & 73.19 & 34.06 & 31.99 & 54.45 \\

\midrule
\multicolumn{7}{c}{\textit{(b) Reward Model Ensemble}}\\
\cmidrule{1-7}

Human Preference
& 79.83 & 60.97 & 71.35 & 20.68 & 30.53 & 52.87 \\
\textit{+} Object Grounding
& 80.44 & 60.01 & 73.79 & 25.84 & 31.15 & 54.03 \\

\textit{++} VQA Consistency
& 80.87 & 59.29 & 74.26 & 30.05 & 31.41 & 53.64 \\
\textit{+++} LMM Alignment
& 82.49 & 61.28 & 73.19 & 34.06 & 31.99 & 54.45 \\
\bottomrule
\end{tabular}
\end{adjustbox}
\end{table}

\begin{table*}[tp]
\centering
\caption{{\bf Performance Comparison on T2I-CompBench++~\cite{huang2025t2i}.} 
The \colorbox{orange!30}{best} and the \colorbox{paleOrange!85}{second-best} scores are highlighted. }
\label{benchmark:t2icomp}
\begin{adjustbox}{width=0.95\linewidth}
\begin{tabular}{l@{\hspace{0.5cm}}cccccccc}
\toprule
\multicolumn{1}{c}{\multirow{2}{*}{\bf Model}} 
& \multicolumn{3}{c}{\bf Attribute Binding } 
& \multicolumn{3}{c}{\bf Object Relationship} 
& \multirow{2}{*}{\bf Numeracy$\uparrow$}
& \multirow{2}{*}{\bf Complex$\uparrow$}
\\
\cmidrule(lr){2-4}\cmidrule(lr){5-7}
& {\bf Color$\uparrow$} 
& {\bf Shape$\uparrow$} 
& {\bf Texture$\uparrow$} 
& {\bf 2D-Spatial$\uparrow$} 
& {\bf 3D-Spatial$\uparrow$} 
& {\bf Non-Spatial$\uparrow$} 
& 
& 
\\
\cmidrule{1-9}
\multicolumn{9}{c}{\textit{Current Generative Models}}\\
\cmidrule{1-9}

Show-o~\cite{xie2024show} 
& 56 & 41 & 46 & 20 & - & 30 & - & 29 \\

SD-XL-base-1.0 \cite{podell2023sdxl} 
& 58.79 & 46.87 & 52.99 & 21.31 & 35.66 & 31.19 & 49.91 & 32.37  \\

Attend-and-Excite \cite{chefer2023attendandexcite}  
& 64.00 & 45.17 & 59.63 & 14.55 & 32.22 & 31.09 & 47.73 & 34.01    \\

PixArt-$\alpha$ \cite{chen2023pixartalpha} 
& 66.90 & 49.27 & 64.77 & 20.64 & - & \colorbox{paleOrange!85}{31.97} & -  & 34.33 \\

GoT~\cite{fang2025got} &65.51 & 50.08 & 58.36 & 24.57 & & 31.13 & -  &37.54 \\

Show-o + PARM~\cite{guo2025can} 
& 75 & 56 & 66 & {29} & - & 31 & - & 37\\

FLUX.1 \cite{flux2024} 
& 74.07 & {57.18} & 69.22 & 28.63 & \colorbox{paleOrange!85}{38.66} & 31.27 & \colorbox{paleOrange!85}{61.85} & 37.03 \\

Emu3~\cite{wang2024emu3} 
& 75.44 & 57.06 & 71.64 & - &-  & - & - & - \\

T2I-R1~\cite{jiang2025t2i} 
& \colorbox{paleOrange!85}{81.30} & \colorbox{paleOrange!85}{58.52} & \colorbox{paleOrange!85}{72.43} & \colorbox{paleOrange!85}{33.78} &-  & 30.90 &60.97 & 39.93 \\

\cmidrule{1-9}
\multicolumn{9}{c}{\textit{Thinking-while-Generating}}\\
\cmidrule{1-9}
Janus-Pro-7B~\cite{chen2025janus} (Baseline)
& 63.59 & 35.28 & 49.36 & 20.61 &32.94  & 30.85 &41.32 & 35.59 \\

\textsc{TwiG-ZS}  & 73.11 & 41.55 & 64.77 & 21.98 & 33.68 & 30.90 & 36.58 & 48.16\\
\textsc{TwiG-SFT} & 74.58 & 52.42 & 67.95 & 27.02 & 35.57 & 31.24 & 51.70 & \colorbox{paleOrange!85}{53.41} \\
\textsc{TwiG-RL}  & \colorbox{orange!30}{82.49} & \colorbox{orange!30}{61.28} & \colorbox{orange!30}{73.19} & \colorbox{orange!30}{34.06} & \colorbox{orange!30}{38.87} & \colorbox{orange!30}{31.99} & \colorbox{orange!30}{61.93}  & \colorbox{orange!30}{53.56}  \\
\bottomrule
\vspace{0.1cm}
\end{tabular}
\end{adjustbox}
\end{table*}

\paragraph{\textsc{TwiG}-50K Dataset.}
To support the task formulation, we curate a high-quality dataset termed \textsc{TwiG}-50K. The construction process comprises multiple stages of synthetic supervision using advanced commercial models.\vspace{0.1cm}
\begin{itemize}
    \item For \textit{what} to say (\(\sim\)17K, three tasks), we source 5.5K text prompts from the training split of T2I-CompBench~\cite{huang2023t2i}, and adopt GPT-4o~\cite{hurst2024gpt} to generate stepwise sub-captions that segment the image into three coherent parts (upper background, central content, lower background). These sub-captions are concatenated and fed to GPT-4o-Image~\cite{hurst2024gpt} to synthesize images that are semantically consistent with the specified divisions. We then filter low-quality instances and organize them into interleaved formats aligned with the \emph{Thinking-while-Generating} protocol. Note that, since the reasoning step count is fixed to three, we do not collect supervision data for \textit{when} to think.\vspace{0.1cm}

    \item For \textit{how} to refine (\(\sim\)17K, three tasks), building on the interleaved samples above, we construct three visual understanding tasks focused on critique and revision. GPT-4o is prompted to evaluate each region by assigning a critic score along five criteria (the same as zero-shot settings) and to provide a revised sub-caption that addresses deficiencies identified by the critique. If the original image attains a high score, the revised thought simply repeats, a case that may not trigger re-generation during inference.\vspace{0.1cm}

    \item To enhance the generation capability of \(\mathrm{ULM}_{g}\) (\(\sim\)16K, three tasks), we construct interleaved visual generation data from the image–sub-caption pairs obtained in the \textit{when/what} stage. Each training instance conditions the generation of region \(\mathcal{V}_k\) on cumulative reasoning thoughts \(\{\tau_j\}_{j \le k}\) and previously generated visual contents \(\{\mathcal{V}_j\}_{j < k}\). Note that this remains text-to-image supervision to preserve a single generation trajectory (not image-to-image), augmented with a visual pre-context.\vspace{-0.2cm}
\end{itemize}

\paragraph{Experiments and Analysis.}
In Table~\ref{tab:sft_mix_nexttosetting} (top), we present the performance of our fine-tuned model, \textsc{TwiG-SFT}. Relative to the zero-shot baseline (\textsc{TwiG-ZS}), SFT delivers modest and reliable gains across benchmarks, with the largest improvements on \emph{Shape} and \emph{Spatial} categories. This demonstrates the effectiveness of our fine-tuning recipe and the curated \textsc{TwiG}-50K dataset. By default, we inherit the optimal model settings from \textsc{TwiG-ZS}, and adopt a balanced data mixture with equal thinking and generation tasks. We further provide two analyses:\vspace{0.1cm}
\begin{itemize}
    \item \textbf{Ablation \textit{(a)}:} Effect of data composition from \textsc{TwiG}-50K. Balancing thinking (\textcolor{G_teal!70!white}{\textbf{\textit{T}}}) and generation (\textcolor{T_yellow!70!white}{\textbf{\textit{G}}}) provides the best trade-off and strengthens \emph{Thinking-while-Generating} from both sides. However, adding reflection data (\textcolor{R_purple!70!white}{\textbf{\textit{R}}}) degrades the results, where the thoughts become longer and over-corrections appear more frequently. This suggests that, \textsc{TwiG-ZS} already exposes most of the model’s reflection proficiency, and oversupplying \textcolor{R_purple!70!white}{\textbf{\textit{R}}} diverts capacity away from learning stable \textcolor{G_teal!70!white}{\textbf{\textit{T}}} and \textcolor{T_yellow!70!white}{\textbf{\textit{G}}} behaviors. Although the reflection subset cannot contribute here, we hope it will facilitate future research on critique-and-revise training.\vspace{0.1cm}

    \item \textbf{Comparison \textit{(b)}:} Inference stability across five random seeds. We report the standard deviation (\textit{Std}) over different runs, and observe that SFT notably tightens dispersion compared to \textsc{TwiG-ZS}, indicating more predictable behavior. Qualitatively, SFT shortens verbose thoughts, curbs hallucinations, improves attribute persistence across adjacent regions, and reduces spurious reflection triggers near the decision threshold.
\end{itemize}

\subsection{Reinforcement Learning}
\label{s4.3}
\vspace{0.1cm}

\paragraph{\textsc{TwiG}-GRPO Strategy.}
To further advance performance, we employ RL to enhance the interleaved reasoning.
Specifically, we adopt the GRPO algorithm~\cite{shao2024deepseekmath_grpo} with the training prompts from T2I-CompBench, and tailor it to our \textit{Thinking-while-Generating} framework. Within this setup, the ULM performs multiple forward passes within a single rollout during GRPO training. A key design question is which components should be reinforced through the reward mechanism: all stages, or only the understanding or generation phases? We propose to reinforce all of them simultaneously through our \textsc{TwiG}-GRPO strategy. Concretely, we compute a single reward based on the final generated image and the input prompt, and utilize it as a shared reward to optimize the policies of every thinking, generation, and reflection pass jointly. This approach not only simplifies implementation (no need to compute rewards for each local visual subtask), but also enables consistent reinforcement across $\mathrm{ULM}_u$ and $\mathrm{ULM}_g$, allowing global information to flow across different paths and thereby enhancing the overall synergy of the \textsc{TwiG} framework.\vspace{-0.2cm}

\begin{figure*}[t]
\vspace{-0.3cm}
    \includegraphics[width=\linewidth]{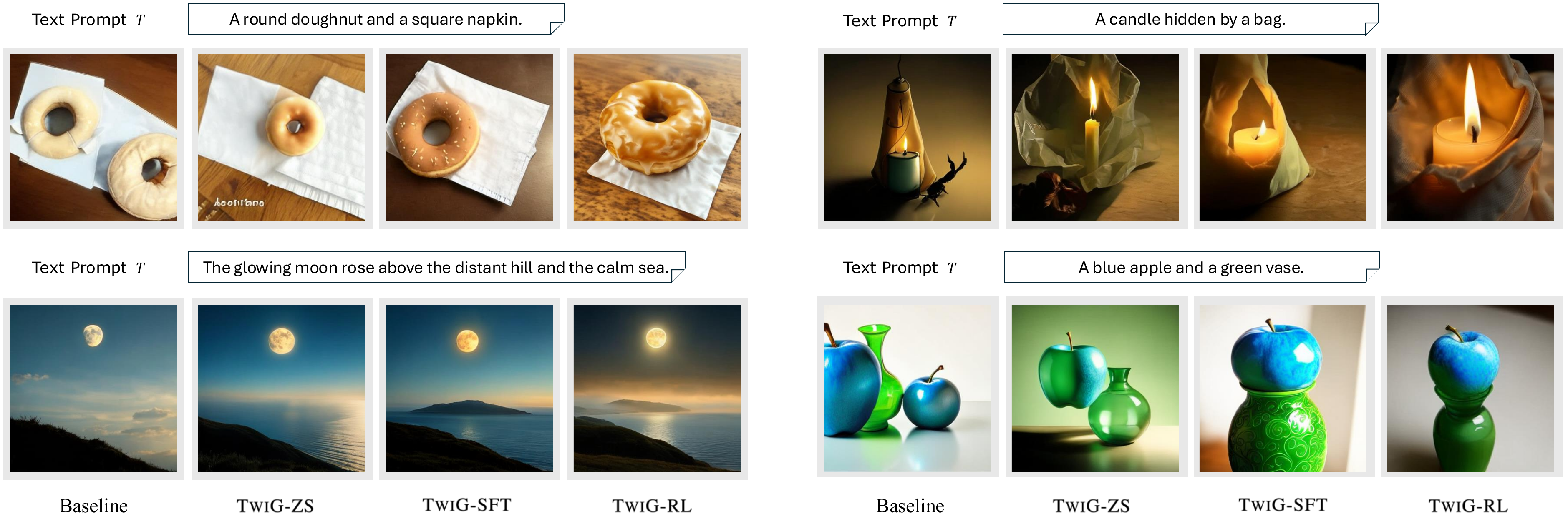}
    \caption{\textbf{Qualitative Comparison of \textsc{TwiG} Variants:} the baseline (Janus-Pro-7B~\cite{chen2025janus}), \textsc{TwiG-ZS}, \textsc{-SFT}, and \textsc{-RL}. Our method demonstrates progressive improvements in compositional fidelity, object counting, and visual realism.}
    \label{fig:compare_vis}
    \vspace{0.2cm}
\end{figure*}

\begin{figure*}[t]
    \includegraphics[width=\linewidth]{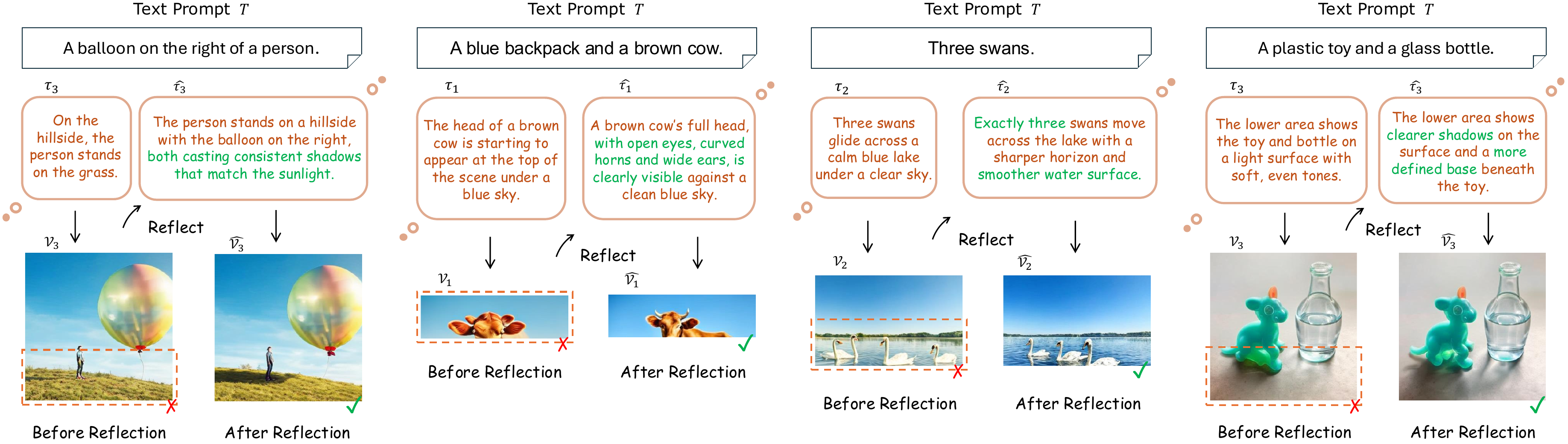}
    \caption{\textbf{The Reflection Capacity of \textsc{TwiG-RL}}. The reflection within our \textit{Thinking-while-Generating} refines both semantic and visual consistency, e.g., improving spatial alignment, shadow coherence, and overall realism across diverse prompts.}
    \label{fig:reflect_vis}
\end{figure*}

\begin{figure*}[t]
\centering
    \includegraphics[width=0.96\linewidth]{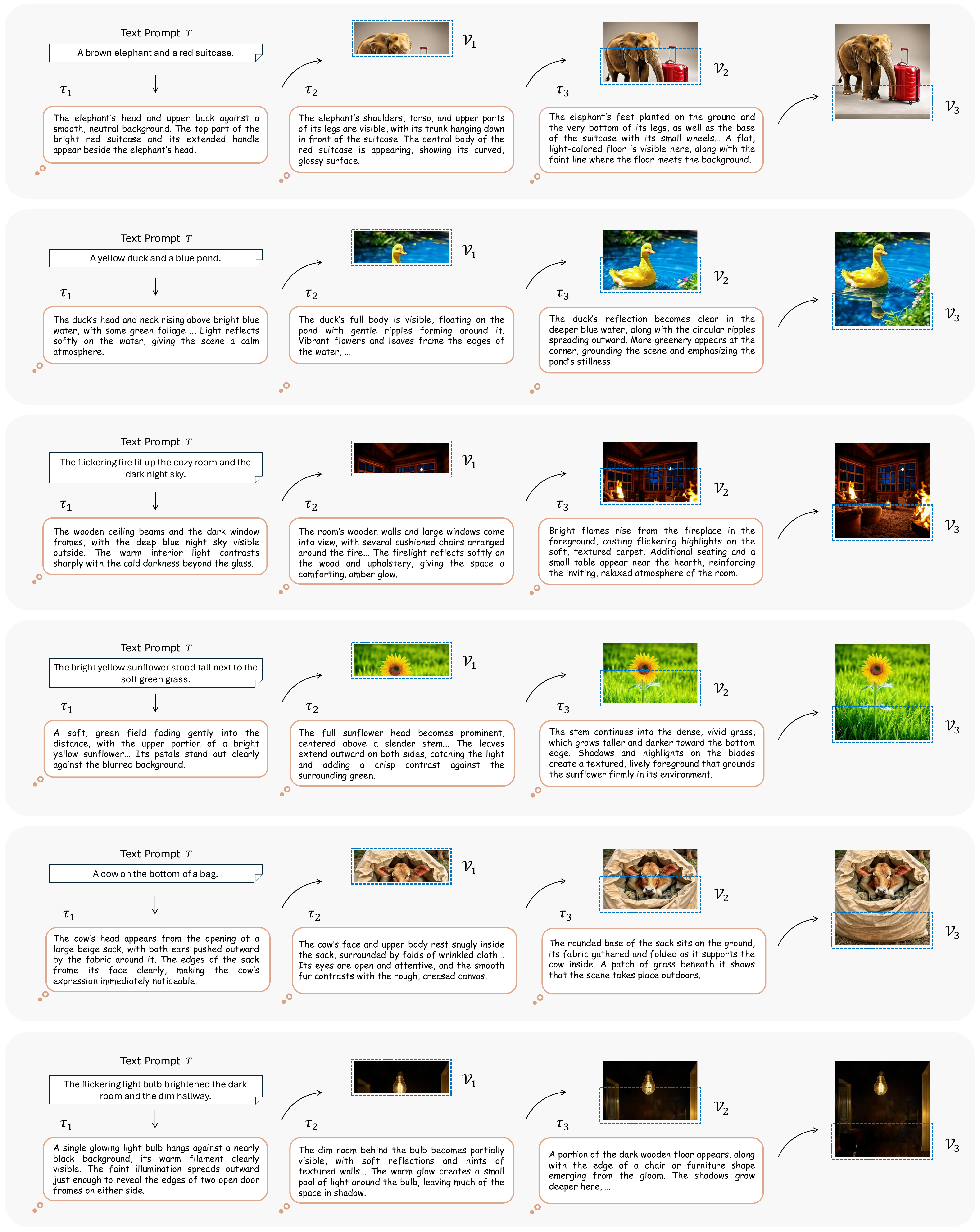}
    \caption{\textbf{\textit{Thinking-while-Generating} Process of \textsc{TwiG-RL.}} Each example showcases how the model iteratively interleaves its textual reasoning and visual outputs, progressively improving compositional accuracy, spatial alignment, and scene coherence.}
    \label{fig:supp_vis}
\end{figure*}

\paragraph{Reward Model Design.}
Since a high-quality image must satisfy multiple aspects (overall aesthetics, object attributes and relationships), we explore to combine complementary reward models for joint optimization and mitigating reward hacking~\cite{jiang2025t2i}: (i) human preference scores (HPS v2~\cite{wu2023human}), (ii) object grounding scores (GroundingDINO~\cite{Liu2023GroundingDM}), (iii) VQA consistency scores (GIT~\cite{wang2022git}), and (iv) LMM alignment scores (the fine-tuned ORM~\cite{guo2025can}). We utilize an unweighted average of the four reward model, and this simple strategy effectively leverages our framework’s generality for RL gains.\vspace{-0.2cm}

\paragraph{Experiments and Analysis.}
In Table~\ref{tab:rl_main_ablation} (top), we present the performance of our reinforced model, \textsc{TwiG-RL}. Compared with \textsc{TwiG-SFT}, the initialization point, RL delivers substantial gains, e.g., exceeding +5\%, across the three \emph{Attribute Binding} categories and the \emph{Spatial} category. This highlights the remaining headroom of the \emph{Think-while-Generating} paradigm once a policy is guided in a right direction with an appropriate GRPO strategy and reward ensemble designs. 
In Table~\ref{benchmark:t2icomp}, we report the three \textsc{TwiG} approaches in comparison with current generative models on T2I-CompBench++~\cite{huang2025t2i}.
Our method offers a flexible trade-off between implementation efficiency (ZS) and competitive performance (RL), allowing practitioners to balance the cost and quality according to deployment needs. Furthermore, in Figures~\ref{fig:compare_vis},~\ref{fig:reflect_vis}, and~\ref{fig:supp_vis}, we present three visualizations, i.e., illustrating the improvements across different variants, the reflection capability, and the image-text interleaved reasoning process, respectively, which highlight the qualitative effectiveness of our methods.\vspace{0.2cm}

\begin{itemize}
    \item \textbf{Ablation \textit{(a)}:} Different strategies for GRPO algorithms.
    Our \textsc{TwiG}-GRPO jointly reinforces all (up to nine) local visual subtasks within a single rollout. We investigate to separately optimize the understanding-related tasks (thinking and reflection) and the generation-related tasks, each using the shared reward to update $\mathrm{ULM}_u$ and $\mathrm{ULM}_g$, respectively.
    As compared, the separate enhancements fail to surpass the joint strategy, highlighting their complementary nature and mutual reinforcement. Only when combined under the full \textsc{TwiG}-GRPO strategy can the RL potential of the interleaved reasoning be fully realized.\vspace{0.1cm}

    \item \textbf{Ablation \textit{(b)}:} Ensemble of multiple reward models.
    We begin with a single HPS v2, and progressively incorporate other three rewards.
    HPS v2 primarily improves global aesthetics and stylistic coherence; GroundingDINO tightens entity presence and localization; GIT curbs instruction violations and strengthens attribute consistency; the fine-tuned ORM improves holistic text–image alignment. 
    Adding components steadily improves performance, and the ensemble of four achieves the best overall balance.
    
\end{itemize}

\section{Conclusion}
\label{sec:conclusion}
In this paper, we introduce the \emph{Thinking-while-Generating} (\textsc{TwiG}) paradigm, an interleaved framework that keeps textual reasoning in the loop during visual generation. Starting from carefully designed zero-shot prompts, then enhancing with SFT, and finally optimizing a policy via RL, our \textsc{TwiG} model learns to think, generate, and reflect within a single visual generation trajectory. We hope this paradigm may inspire future research to fully investigate the potential of interleaved visual generation schemes. 

\paragraph{\textit{Limitations.}}
Given the incapacity of current ULMs, our `\textit{when} to think' utilizes a fixed three-step schedule, which is general but not optimal. As more capable models emerge, learning fully adaptive schedules is a promising next step.
Second, our RL setup employs the original GRPO, already strong, but may be further enhanced by recent variants~\cite{zheng2025group,yu2025dapo}.
Finally, extending \textsc{TwiG} to video, 3D, or image-to-image tasks presents another compelling avenue.

{
    \small
    \bibliographystyle{ieeenat_fullname}
    \bibliography{main}
}

\end{document}